%% file: coin2012_arxiv.tex
\documentclass{article} 
\usepackage{amssymb,amsmath,multirow,rotating}
\usepackage{algorithm,color}
\usepackage{algorithmic}
\usepackage{graphicx}
\usepackage{setspace}
\usepackage{longtable,bbm,natbib}

\newcommand{\IH}{\textit{IH}} 

\title{A Comparative Evaluation of Curriculum Learning with Filtering and Boosting}

\title{An Extensive Evaluation of Filtering Misclassified Instances in Supervised Classification Tasks}
\author{Michael R. Smith and Tony Martinez\\
\small Department of Computer Science, Brigham Young University, Provo, UT 84602 USA\\
\small \texttt{msmith@axon.cs.byu.edu, martinez@cs.byu.edu}}
\date{}
\begin{document}
\maketitle

\begin{abstract}
Not all instances in a data set are equally beneficial for inferring a model of the data.
Some instances (such as outliers) are detrimental to inferring a model of the data.
Several machine learning techniques treat instances in a data set differently during training such as curriculum learning, filtering, and boosting.
However, an automated method for determining how beneficial an instance is for inferring a model of the data does not exist.
In this paper, we present an automated method that orders the instances in a data set by complexity based on the their likelihood of being misclassified (\textit{instance hardness}).
The underlying assumption of this method is that instances with a high likelihood of being misclassified represent more complex concepts in a data set.
Ordering the instances in a data set allows a learning algorithm to focus on the most beneficial instances and ignore the detrimental ones.
We compare ordering the instances in a data set in curriculum learning, filtering and boosting.
We find that ordering the instances significantly increases classification accuracy and that filtering has the largest impact on classification accuracy.
On a set of 52 data sets, ordering the instances increases the average accuracy from 81\% to 84\%.
\end{abstract}

\section{Introduction}
The goal of supervised machine learning is to model a task by ``teaching'' a learning algorithm through the presentation of labeled instances from a data set.
The training instances are generally presented to a learning algorithm in no particular order and are generally treated as being equally important for inferring a model of the data.
This can be problematic for many machine learning algorithms in deciding which initial search direction will lead to the optimal solution.
Without guidance, the learning algorithm may choose an inappropriate initial direction to search the hypothesis space from which it may never be able to fully correct due to an inability to unlearn previously learned concepts.

\begin{figure}[t]
\begin{center}
\input{Diagram1.tex} 
\caption{A hypothetical 2-dimensional data set.}
\label{figure:dataset}
\end{center}
\end{figure}
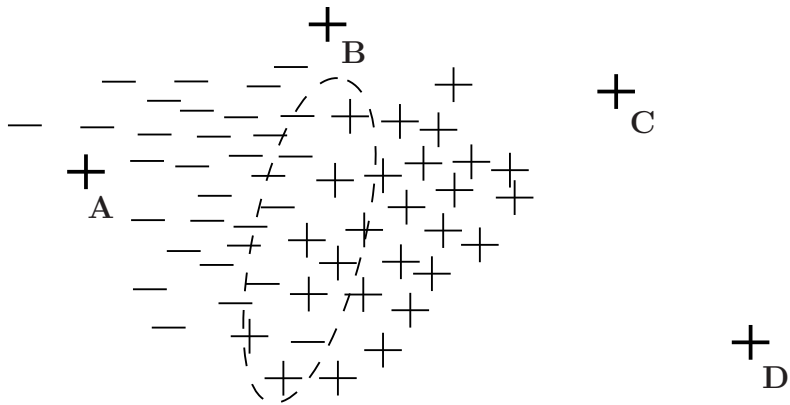

Consider the hypothetical two-dimensional data set in Figure \ref{figure:dataset}.
Instances A, B, C, and D could be considered outliers and represent differing degrees of their extent of being an outlier.
The instances in the dotted oval represent the border points, which could be used to define the classification boundary.
Many learning algorithms have a built in mechanism to avoid overfitting the outliers, however, the presence of outliers could still affect the inferred classification border.
Knowing before training begins which instances are the most informative instances could improve learning.

A number of methods have been developed that treat individual instances in a dataset differently during training to focus on the most informative ones.
Filtering identifies and removes noisy instances and outliers from a data set prior to training and generally results in an increase in classification accuracy on non-filtered test data \citep{Brodley1999,Gamberger2000,Smith2011}.
Boosting also treats instances differently during training by incrementally adjusting the weights of the instances during training \citep{Schapire1990, Freund1995}.
Boosting iteratively trains $m$ base learners and reweights the training data after each model is inferred such that the probability for selecting instances that are misclassified increases.
Boosting, however, has been shown to be prone to overfitting outliers and noisy instances.

\textit{Curriculum learning} was recently formalized by \citet{Bengio2009} as a means of using an ordering of the training data from simplest to most complex to train a learning algorithm.
Instances representing simple concepts are given a weight of 1 while instances that represent complex concepts are initially given a weight of 0, similar to filtering.
As training progresses and the simper training instances are learned, a subset of the more complex training instances receive a weight of 1.
This process continues until all of the training instances receive a weight of 1.
From a cognitive point of view, curriculum learning is based on how humans acquire knowledge.
For example, in schools, subject matter is organized into curricula such that simpler or foundational ideas are presented first.
As learning progresses, more complex concepts and ideas can be learned by using the already learned simpler ideas.
The main deficiencies of most previous work in curriculum learning are: 1) that there is no general method for ordering the instances by complexity and 2) that there is no method for determining when to add more complex instances to the training set.
In previous work for curriculum learning, the ordering was done by hand or by some heuristic specific to the learning task (e.g., the number of words in a sentence).

In this paper, we present an automated method for ordering the instances in a data set based on their likelihood of being misclassified (\textit{instance hardness} \citep{Smith2012_IH}) which we call a \textit{hardness ordering}.
The underlying assumption of ordering the instances by their hardness is that harder instances represent more complex and/or outlier instances.
We use a hardness ordering to examine curriculum learning, filtering, and avoiding overfitting in boosting on a set of 52 UCI data sets using multilayer perceptrons (MLPs) and decision trees (DTs).
As filtering and boosting have received considerable attention \citep{Krause2004,Liu2004}, we focus primarily on developing curriculum learning and comparing curriculum learning with filtering and boosting.
We also examine when to add more complex instances during the training process in curriculum learning.

We find that ordering the instances in a data set by complexity significantly improves classification accuracy.
Specifically, curriculum learning significantly increases classification accuracy for MLPs and significantly \textit{decreases} the classification accuracy for DTs.
Curriculum learning is better suited to learning algorithms that can be incrementally updated similar to MLPs.
Filtering the most complex instances achieves higher classification accuracy than curriculum learning and boosting.
We also examine boosting and curriculum learning with filtering.
Boosting and curriculum learning with filtering significantly increases the accuracy over boosting and curriculum learning without filtering.
Filtering has the most significant effect on the accuracy.
We postulate that the significant change in accuracy when filtering is due to the fact that filtering creates a simpler surface for a learning algorithm to model by removing outliers and noisy instances.
This in turn increases the generalization accuracy.
By contrast, curriculum learning assumes that previously unlearnable complex instances become learnable once the simpler and foundational instances are learned.
By adding the more complex instances into the training process, curriculum learning generates more complex models and the generalization accuracy decreases.

The contributions of this paper include: 1) an automated method for ordering the instances in a data set based on their likelihood of being misclassified,
2) an examination of curriculum learning in general machine learning problems, whereas, previously, curriculum learning has only been examined in limited situations, and
3) a comparison of curriculum learning, filtering and boosting.

The remainder of the paper is organized as follows:
In Section \ref{section:relatedWorks}, we review related works.
In Section \ref{section:curricLearning}, we present how to generate a hardness ordering.
Implementation details of how we implement curriculum learning using a hardness ordering and a comparison with filtering and boosting are presented in Section \ref{section:results}.
Section \ref{section:conclusion} gives conclusions and directions for future work.

\section{Related Works}
\label{section:relatedWorks}
Emphasizing the most important instances has led to success in a number of previous works.
Our work is motivated by the lack of a method to order the instances in curriculum learning and the importance of the early stages of training a learning algorithm.
\citet{Elman1993} realized that the early stages of training dictate what solutions are possible in multilayer perceptrons.
This is true for most gradient descent or greedy learning algorithms.
During the early stages of training, the initial direction to search the hypothesis space is chosen.
All practical machine learning algorithms avoid the computational cost of storing all possible hypotheses consistent with the training data by choosing a promising hypothesis at step $t$.
The hypothesis space is then searched in a step-wise fashion, meaning that the hypothesis found at step $t+1$ is a continuation of the search of the hypothesis space at step $t$.
It can also be difficult for most machine learning algorithms to backtrack and unlearn learned concepts.
This adds more importance to the initial search direction that is chosen to follow.
To aid in this problem, Elman introduced the idea of ``starting small'' meaning that only simple concepts should be used during the early stages of training.
The simple concepts guide the initial search direction of the learning algorithm and can facilitate learning more complex concepts for some situations (Elman used grammar rules).
Other work has demonstrated the utility of starting small in specific application areas \citep{Sanger1994,Neumann2009,Spitkovsky2010, Tu2011}.
Each approach shares the idea of breaking the learning task into subcomponents and then, starting with the simplest concepts, train by gradually increasing concept complexity.

\citet{Bengio2009} formalized these ideas in \textit{curriculum learning}.
The idea behind curriculum learning is to first optimize a smoothed objective function and gradually reduce the degree of smoothing during the training process.
At a more concrete level, curriculum learning is a weighting scheme for training.
Each training instance is assigned a weight which controls how the instance is used in training.
Initially, the weights on the training instances favor the ``easier'' instances  or those that represent simpler concepts.
As training proceeds, the weights on the training instances are updated such that ``harder'' instances and more complex concepts are introduced into the training set.
This continues until all of the instances in the target training set are uniformly weighted.

\citet{Kumar2010} recently presented self-paced learning for latent variable models, building on the idea of curriculum learning.
In self-paced learning, a set of latent variables are learned that indicate which instances should be included for training.
The choice of which instances are used in training is left to the optimization technique and the number of instances used is controlled with a variable which is annealed to eventually use all of the instances.
They assume that the simpler instances are used first for training, but no explicit ordering on the instances is generated.
The primary shortcomings of previous work in curriculum learning are that there is no automated method that orders the instances in a data set by complexity and that there is no method to determine when to add more complex instances to the training set.

Boosting is another approach that incrementally adjusts the weights of the instances during training \citep{Schapire1990, Freund1995}.
Boosting is an algorithm designed to target misclassified instances during training such that as training continues, uninformative instances that lie within a cluster of instances of the same class are suppressed.
Boosting iteratively trains $m$ base learners on a subset of the training data to form an ensemble.
After an iteration of training, the data is weighted such that the probability of selecting instances that are misclassified increases and the probability of selecting the instances that are classified correctly decreases.
These techniques assume that the misclassified instances are the most informative and should be weighted more.
However, this can lead to overfitting noise and new methods were proposed to ignore suspected outliers and weight the more informative or boundary instances higher \citep{Krause2004}.
Boosting differs from curriculum learning in that it is an ensemble method and the influence of the easier instances decrease as training progresses.

Filtering (removing outlier or noisy instances prior to training) is a similar approach to curriculum learning.
Outliers and noisy instances have been observed to adversely affect an induced model \citep{Smith2011}.
Thus, the goal of filtering is to reduce the effects of outlier or noisy instances by removing them prior to training.
One of the difficulties with outlier and noise identification is that there is no agreed-upon definition of what constitutes an outlier or noise.
As such, a variety of different noise and outlier detection methods exist, such as statistical methods \citep{Barnett1978}, density-based clustering \citep{Breunig2000}, and classification-based methods \citep{John95,Brodley1999}.
These methods have been used to identify and remove outliers prior to training, resulting in higher classification accuracy \citep{Brodley1999,Gamberger2000,Liu2004}.

\section{Ordering the Instances}
\label{section:curricLearning}
Many machine learning techniques could benefit from knowing how beneficial an instance is to inferring a model of the data.
In this paper, we use \textit{instance hardness} \citep{Smith2012_IH} to order the instances by complexity.
Instance hardness posits that each instance in a data set has a hardness property that indicates the likelihood that it will be misclassified.
Instances with high instance hardness are frequently misclassified.
For example, outliers and mislabeled instances are expected to have high instance hardness since a learning algorithm will have to overfit to classify them correctly.
The instance hardness of an instance is primarily dependent on its relationship to the other instances in the data set and, by extension, to the underlying data distribution.
Obviously, some instances are harder for some learning algorithms than for others.
For example, some instances are harder for a linear classifier than for a non-linear classifier because a non-linear classifier is capable of producing more complex decision boundaries.
Ideally, instance hardness should be calculated with respect to the underlying data distribution.
However, as we do not have access to the underlying data distribution, instance hardness is estimated empirically using a set of learning algorithms that have been successfully applied to a wide range of problems and, because of their success, they are commonly used in practice.
Thus, instance hardness is also dependent on the method used to estimate the likelihood of a instance being misclassified.

Formally, instance hardness is the probability that an instance $x_i$ will be misclassified with respect to the underlying data distribution $p(x,y)$ of a task:
\begin{align}
\IH(x_i, y_i) &= 1-p(y_i|x_i, p(x,y))
\end{align}
where $y_i$ is the assigned class label for $x_i$.
The unknown distribution $p(x,y)$ is estimated using an observed training set $t$ from $\mathcal{T}$.
We assume that $t$ is the only available sampling of $p(x,y)$.
The training set $t$ is used in place of $p(x,y)$ proceeding forward.
Instance hardness estimates $p(y_i|x_i, t)$ empirically using a set of learning algorithms.
To be explicit, $p(y_i|x_i, t)$ is estimated as $p(y_i|x_i, t, g)$ which is that probability of $y_i$ given $x_i$ as input to learning algorithm $g$ trained on $t$.
Using a single learning algorithm to calculate the probability of $y_i$ given $x_i$ incorporates the bias of the learning algorithm $g$.
To lessen the bias of a single learning algorithm and to gain insight into the objective function of the task, we use a set of learning algorithms to estimate $p(y_i|x_i,t)$.
Integrating over the set of all learning algorithms $\mathcal{G}$ and weighting each learning algorithm by $p(g)$ instance hardness is:
\begin{align}
\IH(x_i, y_i) &= 1-\int_\mathcal{G} p(g) p(y_i|x_i, t,g) \, dg. \label{eq:integral}
\end{align}
The quality of the estimate of instance hardness depends in large part on the prior distribution of learning algorithms $p(g)$.
The distribution $p(g)$ represents the probability that a particular learning algorithm will be used to infer a model of the task.
Not all learning algorithms are equally likely.
For example, modeling a task with a decision tree is more likely than using a learning algorithm that always predicts the same class.
If all learning algorithms were equally likely, then all instances would have the same instance hardness value under the no free lunch theorem \citep{Wolpert1996}.
In this paper, instance hardness is estimated using learning algorithms that are successfully and commonly used in practice.
We refer to this set as the set of \textit{empirically successful learning algorithms} (ESLAs).
A natural way to approximate $p(g)$ is to weight ESLAs with a non-zero weight and to weight the non-ESLAs with a weight of 0.
Since it is impossible to precisely define the evolving set of ESLAs, it is approximated by selecting a diverse subset of the set of ESLAs ($\mathcal{L}$).
Although we use ESLAs to calculate instance hardness for general machine learning algorithms, other sets of learning algorithms could be used to calculate instance hardness such as those algorithms that are used in text mining or the set of neural network algorithms.

Given the set $\mathcal{L}$ of ESLAs, stochastic integration can be used to approximate the integral from Equation \ref{eq:integral}:
\begin{align}
\IH(x_i, y_i) \approx 1 - \frac{1}{|\mathcal{L}|} \sum_{j=1}^{|\mathcal{L}|} p(y_i|x_i, t, g_j)
\end{align}
where $p(g)$ is approximated as $\frac{1}{|\mathcal{L}|}$.
Since all of the learning algorithms in $\mathcal{L}$ do not produce a probability distribution, we use the indicator function to approximate $p(y_i|x_i, t, g)$.

\begin{figure}[t]
\begin{center}
\input{COD.tex}
\caption{Dendrogram of the considered learning algorithms clustered using unsupervised meta-learning.}
\label{figure:COD}
\end{center}
\end{figure}
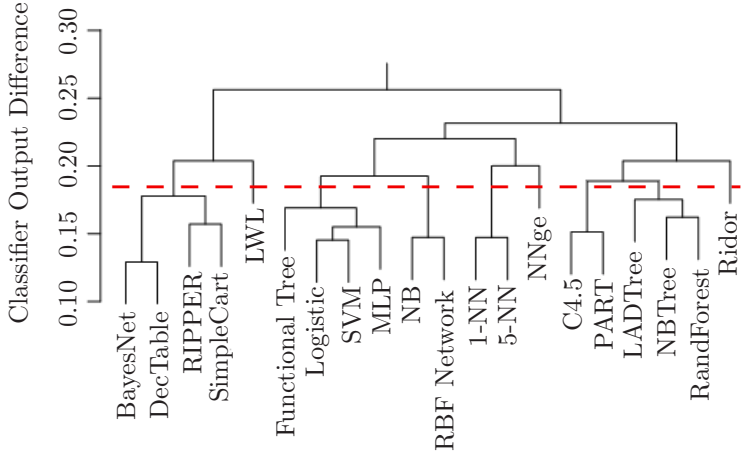

For instance hardness, the diversity of the learning algorithms is determined using unsupervised meta-learning \citep{Lee2011}.
Unsupervised meta-learning uses Classifier Output Difference (COD) \citep{Peterson2005} to measure the diversity between learning algorithms.
COD measures the distance between two learning algorithms as the probability that the learning algorithms make different predictions.
Unsupervised meta-learning then clusters the learning algorithms based on their COD scores with hierarchical agglomerative clustering.
We considered 20 commonly used learning algorithms.
The dendrogram from clustering the considered learning algorithms using unsupervised meta-learning is shown in Figure \ref{figure:COD}.
The height of the line connecting two clusters corresponds to the distance (COD value) between them.
The COD value 0.18 was chosen because the value produced clusters that included learning algorithms from different families of learning algorithms (e.g. decision tree algorithms, nearest-neighbor algorithms, etc.).
A representative algorithm from each cluster was used to create $\mathcal{L}$ as shown in Table \ref{table:LA}.
The actual number of learning algorithms is not as important as selecting a set of learning algorithms that approximately represents the set of ESLAs.
We have found that using a small, diverse set of ESLAs produces significant improvements in accuracy.
\begin{table}[t]
\caption{Set $\mathcal{L}$ of ESLAs used to calculate instance hardness.}
\label{table:LA}
\begin{center}
\begin{tabular}{l }
* RIpple DOwn Rule learner (RIDOR) \\ * Na\"{i}ve Bayes \\
* Multilayer Perceptron trained with Back Propagation \\ * Random Forrest \\
* Locally Weighted Learning (LWL) \\ * 5-nearest neighbors (5NN) \\
* Nearest Neighbor with generalization (NNge) \\ * Decision Tree (C4.5 \citep{Quinlan1993}) \\
* Repeated Incremental Pruning to Produce Error Reduction (RIPPER) \\
\end{tabular}
\end{center}
\end{table}

The learning algorithms are used as implemented in Weka with their default parameters \citep{weka2009}.
By adjusting the parameters, some instances may be correctly classified more consistently.
However, parameter optimization is an expensive process.
Hence, using default parameters gives insight into which instances are misclassified in most practical scenarios.
The learning algorithms are evaluated using 5 by 10-fold cross-validation\footnote{5 by 10-fold cross-validation runs 10-fold cross-validation 5 times, each time with a different random seed for selecting the 10 partitions of the data.}.
We use five folds to better measure the instance hardness of each instance and to protect against the dependency on the data used in each fold.

\section{Empirical Evaluation}
\label{section:results}

With the ordering of the instances provided by instance hardness, we examine how to use a hardness ordering in the learning process.
We examine using a hardness ordering on a set of 52 UCI data sets \citep{UCI2010} shown in Table \ref{table:dataSets}.
We use a hardness ordering in curriculum learning, filtering and boosting.
As curriculum learning is less fully explored, we focus on developing curriculum learning.
We then compare curriculum learning with filtering and boosting.
We implement the methods using multilayer perceptrons (MLPs) trained with backpropagation and decision trees (DTs) trained using C4.5 \citep{Quinlan1993}.
We chose a MLP because the training set can be augmented during training, which is a natural fit for curriculum learning.
Other incremental learning algorithms could also be used.
We train a MLP until convergence, where convergence is determined by a diminishing learning rate.
Initially, the learning rate is set to 0.3 and it is reduced by 70\% if the error on the training data does not decrease over an entire epoch.
Training continues until the learning rate is less than 0.001.
This helps protect against the choice of learning rate.
We also examine curriculum learning in DTs to show that curriculum learning is not appropriate for all learning algorithms and to demonstrate the effects of filtering and boosting in multiple learning algorithms.

\begin{table}[thp]
\centering
\setlength{\tabcolsep}{4.9pt}
\caption{Datasets used organized by number of instances, number of attributes, and attribute type.}
\begin{tabular}{ c | c | c c c }
\hline
\multirow{2}{*}{\# Inst}& \multirow{2}{*}{\# Att} & \multicolumn{3}{| c }{Attribute Type} \\
\cline{3-5} 
 &  & Categorical & Numerical & Mixed \\
\hline
\multirow{3}{*}{\begin{sideways}$<$ 100 \end{sideways}} & $k <$ 10 & Contact Lenses & & Post-Operative \\
\cline{2-5}
& \multirow{2}{*}{$10 < k < 100$} & Lung Cancer &  & Labor \\
& & & & Trains \\
\hline
\multirow{16}{*}{\begin{sideways}$100<M<1000$ \end{sideways}} & \multirow{6}{*}{$k < 10$} & Breast-w & Iris & Teaching- \\
& & Breast Cancer & Ecoli & Assistant \\
& & & Pima Indians &  \\
& & & Glass & \\
& & & Bupa &  \\
& & & Balance Scale &  \\
\cline{2 - 5}
& \multirow{9}{*}{$10 < k < 100$} & Audiology & Ionosphere & Annealing \\
& & Soybean(large) & Wine & Dermatology \\
& & Lymphography & Sonar & Credit-A \\
& & Congressional- & Heart-Statlog & Credit-G \\
& & Voting Records & & Horse Colic \\
& & Vowel & & Heart-c \\
& & Primary-Tumor & & Hepatitis \\
& & Zoo & & Autos \\
& & & & Heart-h \\
\cline{2-5}
& \multirow{1}{*}{$k > 100$} & & & Arrhythmia \\
\hline
\multirow{8}{*}{\begin{sideways}$ 1000 < M < 10000$ \end{sideways}} & \multirow{3}{*}{$k < 10$} & Car Evaluation & Yeast & Abalone \\
& & Chess & & \\
& & Titanic & & \\
\cline{2-5}
& \multirow{5}{*}{$k < 100$} & Splice & Waveform-5000 & Thyroid- \\
& & & Segment & (sick \& \\
& &  & Spambase & hypothyroid) \\
& &  & Ozone level- &  \\
& &  & Detection &  \\
\cline{2-5}
\hline
\multirow{6}{*}{\begin{sideways}$ M > 10000$ \end{sideways}} & \multirow{2}{*}{$k < 10$} & Nursery & MAGIC Gamma- &  \\
 & & & Telescope & \\
\cline{2-5}
 & \multirow{4}{*}{$k < 100$} & Chess- & & Adult-Census- \\
& & (King-Rook vs. &  & Income (KDD) \\
& &  King-Pawn) & & \\
& &  Letter & & \\
\hline
\end{tabular}
\label{table:dataSets}
\end{table}

\subsection{Curriculum Learning}
\setlength{\tabcolsep}{1.5pt}
Curriculum learning trains a learning algorithm on the simplest concepts prior to training on the more complex instances, analogous to teaching a child a subject such as mathematics: addition and subtraction are taught prior to algebra which is taught prior to calculus, etc.
One of the shortcomings of curriculum learning is that there is no automated way to generate curricula for curriculum learning.
Instance hardness is a natural fit for curriculum learning, providing an ordering of the instances from easiest to hardest.

\subsubsection{Implementation Details}
Even with an ordering for curriculum learning, there still remain a couple of questions that need to be addressed for general machine learning problems: 1) the initial complexity of the training instances, and 2) when to add more complex training instances.

The complexity of the initial training instances has a high likelihood of affecting the final model since the initial training instances determine the search direction that will be continued as more instances are added to the training set.
Providing a learning algorithm with instances that are too easy and uninformative could lead to a learning algorithm choosing an arbitrary and/or incorrect gradient to follow.
On the other hand, if the initial complexity is too complex, subconcepts may not be learned since they are grouped with the concepts that build on them.

As training begins, instances with an instance hardness value below a threshold are added to the training set.
To gain insight into curriculum learning, we try different values for the initial complexity of the training set: 0, 0.25, 0.5, and 0.75.
Each time new instances are added to the training set, the instance hardness threshold (\IH) is incremented by 0.1 until all of the training instances are used for training.

When to add more complex instances to a training set could also have an impact on the final outcome of the induced model.
If a learning algorithm does not train on the initial instances long enough, a learning algorithm may not find the optimal gradient to follow.
Training too long on a subset of the instances has the potential for a learning algorithm to overfit the subset.
This could hinder learning the more complex instances that will be introduced later in the training process.

For MLPs, we consider two techniques of when to add more complex instances to the training set.
\begin{description}
 \item[Every $n$ epochs.] For this method, instances with increasing instance hardness values are added to the training set every $n$ epochs. 
 This method is simple and it provides an indication of whether it may be better to let the MLP train more or less before adding more instances to the training set.
 We set $n$ to 25, 50, 100, 200, 300, 400, and 500 to give an indication of how long to train before adding more instances.
 Other values could have been used as well.
 \item[Convergence.] This method trains the MLP to convergence on the training set before adding more complex instances to the training set.
\end{description}

For DTs, more complex instances are added to the training set after a DT is induced using the given training set.
Once more complex instances are added to the training set, the more complex training instances are propagated to a leaf node and the instances are evaluated to determine if the tree should be expanded at this leaf node.
As the training set is augmented, the previously trained portions of the DT are not modified.
For DTs, we consider pruning and not pruning the tree before adding more complex instances to the training set.
For pruning at a given node, the error is estimated for its descendant branches as well as if the node was a leaf node.
If the estimated error of the node as a leaf node is lower that the estimated error of the descendant branches, the node's descendants are pruned.

\subsubsection{Results}

The results from implementing curriculum learning in MLPs and DTs are shown in Table \ref{table:allMLPDSstats}.
Table \ref{table:allMLPDSstats} provides the aggregate statistics for each method of adding more complex instances to the training set (average over all datasets, \textit{p}-value from the Wilcoxon signed-rank test, and number of times that the accuracy from curriculum learning is greater than, equal to, or less than the original).

For MLPs, curriculum learning significantly increases classification accuracy with an alpha value of 0.05 for all of the methods of when to add more complex instances to the training set examined in this work.
Adding more complex instances to the training set after training for 100 epochs is the most significant although adding more more complex instances after 50 training epochs results in the highest average classification accuracy.
The average accuracy increases from 81.41\% to 82.02\% when more complex instances are added to the training set after training for 100 epochs.
It is interesting to note that curriculum learning considerably decreases the classification accuracy on some of the datasets.
The nursery dataset decreases in accuracy from 99.82\% to around 97.42\% regardless of when more complex training instances are added to the training set.
The chess dataset decreases in accuracy as well.
The reasoning behind this could be that there are no subconcepts in the dataset.
On other datasets, curriculum learning increases classification accuracy considerably.
For example, the post-operativePatient dataset increases in accuracy from 68.22\% to 71.11\% and the contact lenses dataset increases from 60\% to 79.17\%.

\begin{table}
\caption{Comparison of different strategies of when to add more complex instances in curriculum learning for MLPs and DTs.}
\label{table:allMLPDSstats}
\begin{center}
\begin{tabular}{ll ccccccccc}
\multicolumn{11}{c}{MLP}\\
\hline
 && Orig  & 25 & 50 & 100 & 200 & 300 & 400 & 500 &  Conv \\
\hline
Average && 81.41  & 82.32 & 82.57 & 82.02 & 82.23 & 82.20 & 81.53 & 82.13 & 82.15 \\
\multicolumn{3}{l}{\textit{p}-values} & \textbf{0.002} & \textbf{0.004} & \textbf{0.001} & \textbf{0.003} & \textbf{0.002} & \textbf{0.002} & \textbf{0.003} & \textbf{0.034} \\
\multicolumn{3}{l}{$>-=-<$} & 36-1-15 & 32-2-18 & 35-2-15 & 34-2-16 & 36-2-14 & 33-2-17 & 33-2-17 & 31-2-19 \\\\
\end{tabular}

\begin{tabular}{ll ccc}
\multicolumn{5}{c}{DT}\\
\hline
  && Orig & Prune & NoPrune \\
\hline
Average & & \textbf{80.12} & 78.62 & 76.56 \\ 
\multicolumn{3}{l}{\textit{p}-values} & 0.998 & 1 \\
\multicolumn{3}{l}{$>-=-<$} & 15-1-36 & 7-0-45\\
\end{tabular}

\end{center}
\end{table}

For DTs, curriculum learning significantly decreases classification accuracy.
Despite this, on a few datasets, curriculum learning considerably increases classification accuracy.
The labor dataset increases from 73.68\%, to 85.26\% and the anneal.ORIG dataset increases from 90.98\% to 94.32\%.
However, other datasets show a considerable decrease in classification accuracy such as the breast-cancer dataset which decreases from 75.52\% to less than 69\%.
Clearly, DTs are not as well suited for curriculum learning as MLPs are.
This may be due to the use of entropy in C4.5 and the inability to backtrack and recover from splitting on a suboptimal attribute.
MLPs, on the other hand, can partially recover from this through weight updates.

One potential problem is that there may not be enough information provided in the initial training set to infer a model of the data.
To test this, we adjusted the instance hardness value of the initial instances in the training set.
(Originally, only instances with an instance hardness of 0 were used in the initial training set).
The aggregate results from setting the instance hardness value of the initial training instances to 0, 0.25, 0.5, and 0.75 are shown in Table \ref{table:easy} for adding more complex instances to the training set after 100 training epochs and training until convergence for MLPs and for pruning and not pruning before adding more complex instances in DTs.
The \textit{p}-values and counts are with respect to the initial training instances having an instance hardness value of 0.

\begin{table}
\caption{Comparison of different initial complexity levels (instance hardness) for the training set.} 
\label{table:easy}
\begin{center}
\begin{tabular}{lllllcccc}
&&& Inital IH: && 0 & 0.25 & 0.5 & 0.75 \\
\hline
\multirow{6}{*}{\begin{sideways}MLP\end{sideways}} && \multirow{3}{*}{\begin{sideways}100\end{sideways}}& Average && 82.02 & 82.17 & \textbf{82.42} & 82.34 \\
&&&\multicolumn{3}{l}{\textit{p}-values} & 0.375 & 0.100 & 0.329 \\
&&&\multicolumn{3}{l}{greater-equal-less} & 26-2-24 & 28-1-23 & 25-2-25 \\
\cline{3-9}
 && \multirow{3}{*}{\begin{sideways}Conv\end{sideways}}&Average && 82.02 & 82.52 & 82.85 & \textbf{83.09} \\
&&&\multicolumn{3}{l}{\textit{p}-values} & 0.107 & \textbf{0.006} & $\mathbf{<0.001}$ \\
&&&\multicolumn{3}{l}{greater-equal-less} & 29-3-20 & 32-3-17 & 36-3-13 \\
\hline
\multirow{6}{*}{\begin{sideways}DT\end{sideways}} && \multirow{3}{*}{\begin{sideways}Prune\end{sideways}}&Average && \textbf{78.62} & \textbf{78.62} & \textbf{78.62} & 79.61 \\
&&&\multicolumn{3}{l}{\textit{p}-values} & 1 & 1 & 0.605 \\
&&&\multicolumn{3}{l}{greater-equal-less} & 0-52-0 & 0-52-0 & 2-29-1 \\
\cline{3-9}
&& \multirow{3}{*}{\begin{sideways}NoP\end{sideways}}&Average && 76.55 & 77.72 & 78.80 & \textbf{79.22} \\
&&&\multicolumn{3}{l}{\textit{p}-values} & $\mathbf{<0.001}$ & $\mathbf{<0.001}$ & $\mathbf{<0.001}$ \\
&&&\multicolumn{3}{l}{greater-equal-less} & 34-3-15 & 43-3-6 & 44-2-6 \\
\hline
\end{tabular}
\end{center}
\end{table}

For MLPs, increasing the instance hardness value of the initial training set significantly increases the classification accuracy with an alpha value of 0.05 when training until convergence before adding more complex instances.
When adding more complex instances to the training set after 100 epochs, increasing the initial complexity did not significantly increase the classification accuracy.
For DTs with pruning, the initial instance hardness value has very little effect--all of the datasets have the same accuracy for initial instance hardness values of 0.25 and 0.5 and only three datasets change in classification accuracy with an initial hardness value of 0.75.
Thus, pruning before adding more complex instances to the training set appears to lessen the affects of the initial instance hardness value.
When DTs are not pruned before adding more complex instance to the training set, an initial instance hardness of 0.75 significantly increases classification accuracy over having an initial instance hardness value of 0.
(It should be remembered that the original classification accuracy for DTs is 80.12\%, thus, any form of curriculum learning for DTs examined so far does not increase the accuracy over the original).

\subsection{Comparison with Filtering and Boosting}
In this section, we compare curriculum learning with filtering and boosting.
The filtering technique employed removes any instance with an instance hardness value greater than or equal to a threshold (we use 0.75) \citep{Smith2012_FS}.
We denote the filtering method as IH$_{.75}$.
Evaluation for filtering is done on unfiltered test data.
We compare curriculum learning with two boosting techniques: AdaBoost \citep{Freund1996} (AB) and MultiBoost \citep{Webb2000} (MB).

For curriculum learning with MLPs, we use the method of adding more complex instances to the training set after training for 100 epochs since it was the most significant.
For DTs, pruning is done before adding more complex instances to the training set since it achieves higher classification accuracy.
A statistical comparison of curriculum learning, filtering, and boosting for MLPs and DTs is given in Tables \ref{MLP_stats} and \ref{DT_stats} respectively.

\begin{table}[t]
\caption{A pair-wise comparison of curriculum learning with filtering and boosting for MLPs.
The first row gives the \textit{p}-values from the Wilcoxon signed-rank test for statistical significance.
The second row gives the number of data sets with greater, equal, or lower classification accuracy.}
\label{MLP_stats}
\begin{center}
\begin{tabular}{lcccccccc}
& Orig & IH$_{.75}$ & AB & MB & CL & AB$_{.75}$ & MB$_{.75}$ & CL$_{.75}$ \\
\hline
Average & 81.41 & \textbf{83.96} & 81.72 & 81.85 & 82.32 & \textbf{84.05} & \textbf{83.87} & \textbf{83.99} \\
\hline
\multirow{2}{*}{Orig} & 1 & 1 & 0.768 & 0.987 & 0.999 & 1 & 1 & 1 \\
 & 0-52-0 & 11-2-39 & 25-2-25 & 20-2-30 & 15-2-35 & 11-0-41 & 11-1-40 & 12-2-38 \\
\hline
\multirow{2}{*}{IH$_{.75}$} & $\mathbf{<0.001}$ & 1 & $\mathbf{<0.001}$ & $\mathbf{<0.001}$ & $\mathbf{<0.001}$ & 0.925 & 0.873 & 0.569 \\
 & 39-2-11 & 0-52-0 & 39-0-13 & 40-0-12 & 39-4-9 & 20-1-31 & 19-1-32 & 23-6-23 \\
\hline
\multirow{2}{*}{AB} & 0.234 & 1 & 1 & 0.981 & 0.996 & 1 & 1 & 1 \\
 & 25-2-25 & 13-0-39 & 0-52-0 & 15-14-23 & 15-0-37 & 9-1-42 & 12-1-39 & 9-1-42 \\
\hline
\multirow{2}{*}{MB} & \textbf{0.013} & 1 & \textbf{0.020} & 1 & 0.986 & 1 & 1 & 1 \\
 & 30-2-20 & 12-0-40 & 23-14-15 & 0-52-0 & 17-0-35 & 7-2-43 & 12-1-39 & 10-1-41 \\
\hline
\multirow{2}{*}{CL} & \textbf{0.001} & 1 & \textbf{0.004} & \textbf{0.014} & 1 & 1 & 1 & 1 \\
 & 35-2-15 & 9-4-39 & 37-0-15 & 35-0-17 & 0-52-0 & 11-1-40 & 13-1-38 & 9-1-42 \\
\hline
\multirow{2}{*}{AB$_{.75}$} & $\mathbf{<0.001}$ & 0.076 & $\mathbf{<0.001}$ & $\mathbf{<0.001}$ & $\mathbf{<0.001}$ & 1 & \textbf{0.028} & 0.211 \\
 & 41-0-11 & 31-1-20 & 42-1-9 & 43-2-7 & 40-1-11 & 0-52-0 & 19-21-12 & 29-1-22 \\
\hline
\multirow{2}{*}{MB$_{.75}$} & $\mathbf{<0.001}$ & 0.129 & $\mathbf{<0.001}$ & $\mathbf{<0.001}$ & $\mathbf{<0.001}$ & 0.973 & 1 & 0.349 \\
 & 40-1-11 & 32-1-19 & 39-1-12 & 39-1-12 & 38-1-13 & 12-21-19 & 0-52-0 & 28-1-23 \\
\hline
\multirow{2}{*}{CL$_{.75}$} & $\mathbf{<0.001}$ & 0.435 & $\mathbf{<0.001}$ & $\mathbf{<0.001}$ & $\mathbf{<0.001}$ & 0.791 & 0.655 & 1 \\
 & 38-2-12 & 23-6-23 & 42-1-9 & 41-1-10 & 42-1-9 & 22-1-29 & 23-1-28 & 0-52-0 \\
\end{tabular}
\end{center}
\end{table}

\begin{table}[t]
\caption{A pair-wise comparison of curriculum learning with filtering and boosting for DTs.
The first row gives the \textit{p}-values from the Wilcoxon signed-rank test for statistical significance.
The second row gives the number of data sets with greater, equal, or lower classification accuracy.}
\label{DT_stats}
\begin{center}
\setlength{\tabcolsep}{1.3pt}
\begin{tabular}{lcccccccc}
& Orig & IH$_{.75}$ & AB & MB & CL & AB$_{.75}$ & MB$_{.75}$ & CL$_{.75}$ \\
\hline
Average & 80.12 & 82.04 & 81.52 & 81.88 & 78.62 & \textbf{84.54} & \textbf{83.84} & 82.16 \\
\hline
\multirow{2}{*}{Orig} & 1 & 1 & 0.998 & 1 & \textbf{0.002} & 1 & 1 & 1 \\
 & 0-52-0 & 8-1-43 & 14-1-37 & 13-0-39 & 36-1-15 & 2-2-48 & 2-3-47 & 10-1-41 \\
\hline
\multirow{2}{*}{IH$_{.75}$} & $\mathbf{<0.001}$ & 1 & 0.332 & 0.595 & $\mathbf{<0.001}$ & 1 & 1 & 0.457 \\
 & 43-1-8 & 0-52-0 & 23-1-28 & 23-0-29 & 43-1-8 & 4-5-43 & 9-2-41 & 26-6-20 \\
\hline
\multirow{2}{*}{AB} & \textbf{0.002} & 0.672 & 1 & 0.957 & $\mathbf{<0.001}$ & 1 & 1.000 & 0.771 \\
 & 37-1-14 & 28-1-23 & 0-52-0 & 21-0-31 & 44-0-8 & 9-2-41 & 15-2-35 & 27-2-23 \\
\hline
\multirow{2}{*}{MB} & $\mathbf{<0.001}$ & 0.408 & \textbf{0.044} & 1 & $\mathbf{<0.001}$ & 1 & 1 & 0.542 \\
 & 39-0-13 & 29-0-23 & 31-0-21 & 0-52-0 & 48-1-3 & 5-1-46 & 12-1-39 & 26-0-26 \\
\hline
\multirow{2}{*}{CL} & 0.998 & 1 & 1 & 1 & 1 & 1 & 1 & 1 \\
 & 15-1-36 & 8-1-43 & 8-0-44 & 3-1-48 & 0-52-0 & 1-0-51 & 2-0-50 & 6-1-45 \\
\hline
\multirow{2}{*}{AB$_{.75}$} & $\mathbf{<0.001}$ & $\mathbf{<0.001}$ & $\mathbf{<0.001}$ & $\mathbf{<0.001}$ & $\mathbf{<0.001}$ & 1 & $\mathbf{<0.001}$ & $\mathbf{<0.001}$ \\
 & 48-2-2 & 43-5-4 & 41-2-9 & 46-1-5 & 51-0-1 & 0-52-0 & 33-5-14 & 43-6-3 \\
\hline
\multirow{2}{*}{MB$_{.75}$} & $\mathbf{<0.001}$ & $\mathbf{<0.001}$ & $\mathbf{<0.001}$ & $\mathbf{<0.001}$ & $\mathbf{<0.001}$ & 1 & 1 & $\mathbf{<0.001}$ \\
 & 47-3-2 & 41-2-9 & 35-2-15 & 39-1-12 & 50-0-2 & 14-5-33 & 0-52-0 & 42-2-8 \\
\hline
\multirow{2}{*}{CL$_{.75}$} & $\mathbf{<0.001}$ & 0.548 & 0.232 & 0.462 & $\mathbf{<0.001}$ & 1 & 1 & 1 \\
 & 41-1-10 & 20-6-26 & 23-2-27 & 26-0-26 & 45-1-6 & 3-6-43 & 8-2-42 & 0-52-0 \\
\end{tabular}
\end{center}
\end{table}

The first row shows the average accuracy for each method.
Each following pair of rows gives the \textit{p}-value from the Wilcoxon signed-rank test and the number of times that a method achieved an accuracy greater than-equal to-less than the method in the column heading.
Curriculum learning (CL) significantly increases classification accuracy over boosting for MLPs.
For both MLPs and DTs, filtering and boosting significantly increases the classification accuracy over the original.

We also examined filtering prior to boosting and curriculum learning.
Curriculum learning with filtering is denoted as CL$_{.75}$ and boosting with filtering is denoted as AB$_{.75}$ and MB$_{.75}$ for AdaBoost and MultiBoost.
For curriculum learning, we set the initial complexity level to 0.5.
When a method is augmented with filtering, the filtered method significantly outperforms (in terms of classification accuracy) the method without filtering.
This is particularly apparent with AdaBoost.
AdaBoost does not significantly increase the classification accuracy over any of the other methods for MLPs and only over the original for DTs.
Yet, with filtering, AB$_{.75}$ significantly increases the classification accuracy over the other methods.
This shows how prone AdaBoost is to overfitting.
These results also show the importance of treating instances differently during training, particularly in the case of AdaBoost and filtering.
Filtering has the most significant effect on accuracy, achieving higher classification accuracy than the original algorithm, curriculum learning, and boosting.
Applying filtering to curriculum learning and boosting also increases the classification accuracy.
AB$_{.75}$ achieves the highest classification accuracy out of all of the investigated methods increasing the accuracy to 84\% for MLPs and DTs.

Concerning curriculum learning, these results lead to the question of what is gained by using curriculum learning and under what circumstances is using curriculum learning most appropriate.
Only 2 of the 52 datasets (autos and labor) have a considerably higher classification accuracy for curriculum learning than filtering for both MLPs and DTs.
Originally, curriculum learning was proposed to be used in specific tasks as a continuation method that were assumed to be non-convex.
However, it is difficult to determine if a data set represents a non-convex problem.
A common approach for determining convexity is taking the difference in accuracy between a linear classifier and a nonlinear classifier.
We used the relative percentage difference between the accuracy of a MLP trained with backpropagation and a perceptron (MLP-per) and between the accuracy of a random forest and a linear SVM (RF-SVM).
Non-convexity is assumed if the non-linear classifier significantly out performs the linear classifier (although this really determines non-linearity).

We compared the convexity measures for each data set with the accuracies from curriculum learning with filtering and boosting.
The convexity measures are inconclusive.
For example, for the autos dataset, curriculum learning provides a boost in accuracy as is expected since both MLP-per and RF-SVM are positive.
The classification accuracy for the autos dataset is considerably higher for CL Prune (using a DT), increasing almost 5 percent
On the other hand, a boost is expected on the colic dataset, yet the classification accuracy decreased from about 86\% to 82\%.
Non-convexity, therefore, is not a sufficient condition for curriculum learning to outperform filtering the dataset.
It is difficult to know when to use curriculum learning for general machine learning tasks, especially when a simple method such as filtering produces similar results.

\section{Conclusions} 
\label{section:conclusion}
In this paper we presented a method for ordering the instances in a data set by complexity (hardness ordering).
A hardness ordering uses instance hardness to order the instances in a data set based on the their likelihood of being misclassified.
The hardness ordering allows a learning algorithm to focus on the most informative instances.

We integrated the hardness ordering for a data set into the learning process in curriculum learning.
One of the main shortcomings of curriculum learning is the lack of a method for developing curricula.
As curriculum learning is a relatively new approach, we examined using a hardness ordering as a general approach to implement curriculum learning.
We examined curriculum learning on a set of 52 UCI data sets using MLPs and DTs and compared curriculum learning with filtering and boosting.

Our exploration with curriculum learning has shed interesting, and unexpected, light that curriculum learning performs strikingly similar to filtering and boosting in MLPs.
The similarity of curriculum learning in MLPs to filtering is somewhat expected.
\citet{Elman1993} pointed out that one of the reasons starting small is so important is due to backpropagation's inflexibility of learning late in the learning process.
As training progresses, weights become rigid and only very small changes are made during training.
As complex instances are not trained on in curriculum learning until late in the learning process, they will only have a very minor (if any) affect on the trained model, especially with a large number of training epochs between adding more instances.
Thus, the harder instances are not expected to have a large impact on the final model.

For the DTs, the hope was that by starting with easier instances, more appropriate attribute splits would be chosen early in the learning process.
As more difficult (and possibly noisy) instances are added later they would have less of an effect on the tree.
DTs may not be an ideal candidate for curriculum learning because they are somewhat robust to noise due to using entropy to choose which attribute to split on and because pruning helps avoid overfitting the data.

The claim that curriculum learning is well suited for non-convex problems may or may not be true.
One difficult aspect of this claim is determining the convexity of a dataset.
Curriculum learning has an intuitive motivation and has proven to work well in specific past applications.
For general machine learning problems, however, it would appear that the benefits may not be as great as originally intended.
Curriculum learning may be well suited for tasks that can be broken down into subtasks.

Our finding that curriculum learning does not out perform filtering and boosting in general machine learning problems may be an artifact of how we ordered the instances in a data set.
Using instance hardness to develop curricula orders the instances based on how hard they are to correctly classify but may not take into account if some concepts are sub-concepts, how concepts are related, etc.
A better understanding of how the concepts in a data set are related could aid in creating more appropriate curricula for a data set.
Future work for curriculum learning includes better understanding of how the instances in a data set are related to each other and developing curricula that accounts for the concepts contained in the instances.

We showed the positive impact of filtering instances prior to training.
Filtering the noisy and outlier instances prior to training significantly increases the accuracy for the original learning algorithm as well as for curriculum learning and boosting.
In fact, filtering prior to using AdaBoost produce significantly higher classification accuracy than all other methods investigated for MLPs and DTs.
By filtering the noisy and outlier instances prior to training, AdaBoost does not overfit the noise since it is no longer in the training data.
AdaBoost further increases classification accuracy by focusing on the instances that are hardest to correctly classify.

\bibliographystyle{apa}
\bibliography{../../bibliography}

\end{document}

%% file: Diagram1.tex
\begingroup
  \makeatletter
  \providecommand\color[2][]{%
    \GenericError{(gnuplot) \space\space\space\@spaces}{%
      Package color not loaded in conjunction with
      terminal option `colourtext'%
    }{See the gnuplot documentation for explanation.%
    }{Either use 'blacktext' in gnuplot or load the package
      color.sty in LaTeX.}%
    \renewcommand\color[2][]{}%
  }%
  \providecommand\includegraphics[2][]{%
    \GenericError{(gnuplot) \space\space\space\@spaces}{%
      Package graphicx or graphics not loaded%
    }{See the gnuplot documentation for explanation.%
    }{The gnuplot epslatex terminal needs graphicx.sty or graphics.sty.}%
    \renewcommand\includegraphics[2][]{}%
  }%
  \providecommand\rotatebox[2]{#2}%
  \@ifundefined{ifGPcolor}{%
    \newif\ifGPcolor
    \GPcolorfalse
  }{}%
  \@ifundefined{ifGPblacktext}{%
    \newif\ifGPblacktext
    \GPblacktexttrue
  }{}%
  \let\gplgaddtomacro\g@addto@macro
  \gdef\gplbacktext{}%
  \gdef\gplfronttext{}%
  \makeatother
  \ifGPblacktext
    \def\colorrgb#1{}%
    \def\colorgray#1{}%
  \else
    \ifGPcolor
      \def\colorrgb#1{\color[rgb]{#1}}%
      \def\colorgray#1{\color[gray]{#1}}%
      \expandafter\def\csname LTw\endcsname{\color{white}}%
      \expandafter\def\csname LTb\endcsname{\color{black}}%
      \expandafter\def\csname LTa\endcsname{\color{black}}%
      \expandafter\def\csname LT0\endcsname{\color[rgb]{1,0,0}}%
      \expandafter\def\csname LT1\endcsname{\color[rgb]{0,1,0}}%
      \expandafter\def\csname LT2\endcsname{\color[rgb]{0,0,1}}%
      \expandafter\def\csname LT3\endcsname{\color[rgb]{1,0,1}}%
      \expandafter\def\csname LT4\endcsname{\color[rgb]{0,1,1}}%
      \expandafter\def\csname LT5\endcsname{\color[rgb]{1,1,0}}%
      \expandafter\def\csname LT6\endcsname{\color[rgb]{0,0,0}}%
      \expandafter\def\csname LT7\endcsname{\color[rgb]{1,0.3,0}}%
      \expandafter\def\csname LT8\endcsname{\color[rgb]{0.5,0.5,0.5}}%
    \else
      \def\colorrgb#1{\color{black}}%
      \def\colorgray#1{\color[gray]{#1}}%
      \expandafter\def\csname LTw\endcsname{\color{white}}%
      \expandafter\def\csname LTb\endcsname{\color{black}}%
      \expandafter\def\csname LTa\endcsname{\color{black}}%
      \expandafter\def\csname LT0\endcsname{\color{black}}%
      \expandafter\def\csname LT1\endcsname{\color{black}}%
      \expandafter\def\csname LT2\endcsname{\color{black}}%
      \expandafter\def\csname LT3\endcsname{\color{black}}%
      \expandafter\def\csname LT4\endcsname{\color{black}}%
      \expandafter\def\csname LT5\endcsname{\color{black}}%
      \expandafter\def\csname LT6\endcsname{\color{black}}%
      \expandafter\def\csname LT7\endcsname{\color{black}}%
      \expandafter\def\csname LT8\endcsname{\color{black}}%
    \fi
  \fi
  \setlength{\unitlength}{0.0500bp}%
  \begin{picture}(6000.00,3000.00)%
    \gplgaddtomacro\gplbacktext{%
      \csname LTb\endcsname%
      \put(810,1444){\makebox(0,0)[r]{\strut{} \textbf{\large A}}}%
      \put(2710,2618){\makebox(0,0)[r]{\strut{} \textbf{\large B}}}%
      \put(4890,2093){\makebox(0,0)[r]{\strut{} \textbf{\large C}}}%
      \put(5900,167){\makebox(0,0)[r]{\strut{} \textbf{\large D}}}%
    }%
    \gplbacktext
    \put(0,0){\includegraphics[scale=0.25]{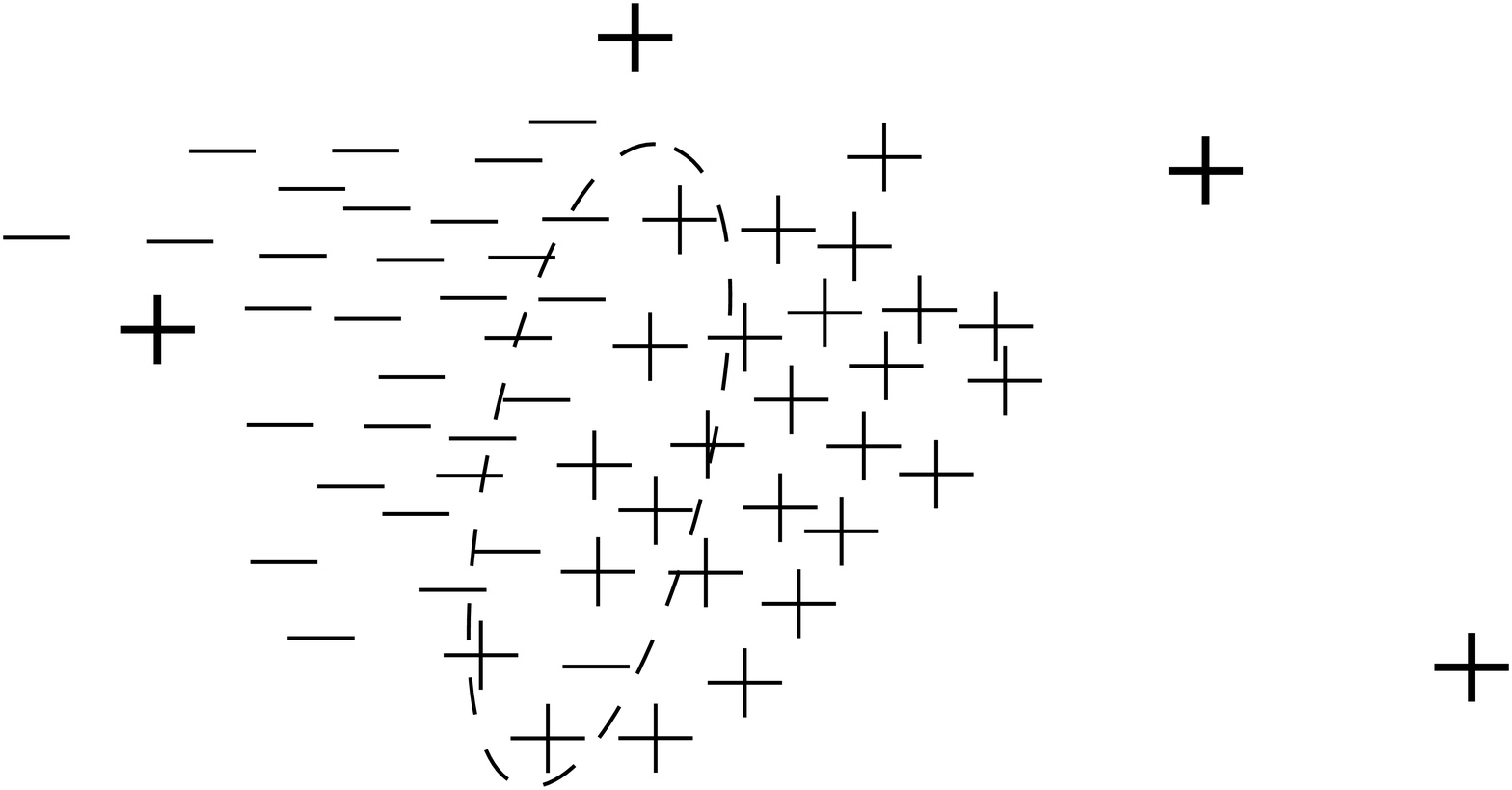}}%
    \gplfronttext
  \end{picture}%
\endgroup

%% file: COD.tex
\begingroup
  \makeatletter
  \providecommand\color[2][]{%
    \GenericError{(gnuplot) \space\space\space\@spaces}{%
      Package color not loaded in conjunction with
      terminal option `colourtext'%
    }{See the gnuplot documentation for explanation.%
    }{Either use 'blacktext' in gnuplot or load the package
      color.sty in LaTeX.}%
    \renewcommand\color[2][]{}%
  }%
  \providecommand\includegraphics[2][]{%
    \GenericError{(gnuplot) \space\space\space\@spaces}{%
      Package graphicx or graphics not loaded%
    }{See the gnuplot documentation for explanation.%
    }{The gnuplot epslatex terminal needs graphicx.sty or graphics.sty.}%
    \renewcommand\includegraphics[2][]{}%
  }%
  \providecommand\rotatebox[2]{#2}%
  \@ifundefined{ifGPcolor}{%
    \newif\ifGPcolor
    \GPcolorfalse
  }{}%
  \@ifundefined{ifGPblacktext}{%
    \newif\ifGPblacktext
    \GPblacktexttrue
  }{}%
  \let\gplgaddtomacro\g@addto@macro
  \gdef\gplbacktext{}%
  \gdef\gplfronttext{}%
  \makeatother
  \ifGPblacktext
    \def\colorrgb#1{}%
    \def\colorgray#1{}%
  \else
    \ifGPcolor
      \def\colorrgb#1{\color[rgb]{#1}}%
      \def\colorgray#1{\color[gray]{#1}}%
      \expandafter\def\csname LTw\endcsname{\color{white}}%
      \expandafter\def\csname LTb\endcsname{\color{black}}%
      \expandafter\def\csname LTa\endcsname{\color{black}}%
      \expandafter\def\csname LT0\endcsname{\color[rgb]{1,0,0}}%
      \expandafter\def\csname LT1\endcsname{\color[rgb]{0,1,0}}%
      \expandafter\def\csname LT2\endcsname{\color[rgb]{0,0,1}}%
      \expandafter\def\csname LT3\endcsname{\color[rgb]{1,0,1}}%
      \expandafter\def\csname LT4\endcsname{\color[rgb]{0,1,1}}%
      \expandafter\def\csname LT5\endcsname{\color[rgb]{1,1,0}}%
      \expandafter\def\csname LT6\endcsname{\color[rgb]{0,0,0}}%
      \expandafter\def\csname LT7\endcsname{\color[rgb]{1,0.3,0}}%
      \expandafter\def\csname LT8\endcsname{\color[rgb]{0.5,0.5,0.5}}%
    \else
      \def\colorrgb#1{\color{black}}%
      \def\colorgray#1{\color[gray]{#1}}%
      \expandafter\def\csname LTw\endcsname{\color{white}}%
      \expandafter\def\csname LTb\endcsname{\color{black}}%
      \expandafter\def\csname LTa\endcsname{\color{black}}%
      \expandafter\def\csname LT0\endcsname{\color{black}}%
      \expandafter\def\csname LT1\endcsname{\color{black}}%
      \expandafter\def\csname LT2\endcsname{\color{black}}%
      \expandafter\def\csname LT3\endcsname{\color{black}}%
      \expandafter\def\csname LT4\endcsname{\color{black}}%
      \expandafter\def\csname LT5\endcsname{\color{black}}%
      \expandafter\def\csname LT6\endcsname{\color{black}}%
      \expandafter\def\csname LT7\endcsname{\color{black}}%
      \expandafter\def\csname LT8\endcsname{\color{black}}%
    \fi
  \fi
  \setlength{\unitlength}{0.0500bp}%
  \begin{picture}(6000.00,3200.00)%
    \gplgaddtomacro\gplfronttext{%
      \csname LTb\endcsname%
      \put(560,1044){\rotatebox{90}{\makebox(0,0)[r]{\strut{} 0.10}}}%
      \put(560,1570){\rotatebox{90}{\makebox(0,0)[r]{\strut{} 0.15}}}%
      \put(560,2093){\rotatebox{90}{\makebox(0,0)[r]{\strut{} 0.20}}}%
      \put(560,2600){\rotatebox{90}{\makebox(0,0)[r]{\strut{} 0.25}}}%
      \put(560,3100){\rotatebox{90}{\makebox(0,0)[r]{\strut{} 0.30}}}%
      \put(1000,812){\rotatebox{90}{\makebox(0,0)[r]{\strut{}BayesNet}}}%
      \put(1250,812){\rotatebox{90}{\makebox(0,0)[r]{\strut{}DecTable}}}%
      \put(1500,1112){\rotatebox{90}{\makebox(0,0)[r]{\strut{}RIPPER}}}%
      \put(1700,1112){\rotatebox{90}{\makebox(0,0)[r]{\strut{}SimpleCart}}}%
      \put(1970,1600){\rotatebox{90}{\makebox(0,0)[r]{\strut{}LWL}}}%
      \put(2200,1232){\rotatebox{90}{\makebox(0,0)[r]{\strut{}Functional Tree}}}%
      \put(2425,1000){\rotatebox{90}{\makebox(0,0)[r]{\strut{}Logistic}}}%
      \put(2700,1012){\rotatebox{90}{\makebox(0,0)[r]{\strut{}SVM}}}%
      \put(2925,1075){\rotatebox{90}{\makebox(0,0)[r]{\strut{}MLP}}}%
      \put(3150,1012){\rotatebox{90}{\makebox(0,0)[r]{\strut{}NB}}}%
      \put(3400,975){\rotatebox{90}{\makebox(0,0)[r]{\strut{}RBF Network}}}%
      \put(3650,1012){\rotatebox{90}{\makebox(0,0)[r]{\strut{}1-NN}}}%
      \put(3875,1012){\rotatebox{90}{\makebox(0,0)[r]{\strut{}5-NN}}}%
      \put(4100,1530){\rotatebox{90}{\makebox(0,0)[r]{\strut{}NNge}}}%
      \put(4365,1062){\rotatebox{90}{\makebox(0,0)[r]{\strut{}C4.5}}}%
      \put(4600,1062){\rotatebox{90}{\makebox(0,0)[r]{\strut{}PART}}}%
      \put(4820,1262){\rotatebox{90}{\makebox(0,0)[r]{\strut{}LADTree}}}%
      \put(5075,1142){\rotatebox{90}{\makebox(0,0)[r]{\strut{}NBTree}}}%
      \put(5325,1142){\rotatebox{90}{\makebox(0,0)[r]{\strut{}RandForest}}}%
      \put(5535,1572){\rotatebox{90}{\makebox(0,0)[r]{\strut{}Ridor}}}%
      \put(190,1930){\rotatebox{90}{\makebox(0,0){\strut{}Classifier Output Difference}}}%
    }%
    \gplbacktext
    \put(0,0){\includegraphics[scale=0.75]{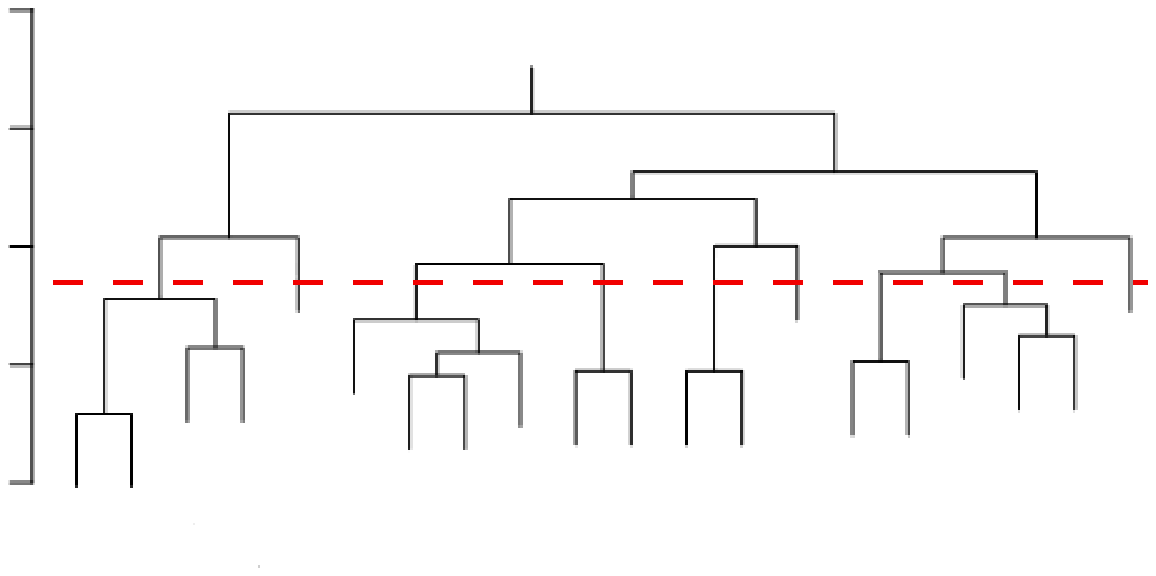}}%
    \gplfronttext
  \end{picture}%
\endgroup